\documentclass[runningheads,a4paper]{llncs}

\usepackage{amssymb}
\usepackage[export]{adjustbox}
\usepackage{amsmath}
\usepackage{graphicx}
\usepackage{bbding}
\usepackage[multi-part-units=single]{siunitx}
\usepackage{url}
\usepackage{multirow}
\usepackage[misc]{ifsym}
\usepackage{bm}

\newcommand{\keywords}[1]{\par\addvspace\baselineskip
\noindent\keywordname\enspace\ignorespaces#1}

\begin{document}

\mainmatter  

\title{Skin Lesion Classification Using Ensembles of Multi-Resolution EfficientNets with Meta Data}

\titlerunning{Skin Lesion Classification in the ISIC2019 Challenge}

%
%

\author{Nils Gessert\inst{12} \and Maximilian Nielsen\inst{23} \and Mohsin Shaikh\inst{23} \and Ren\'e Werner\inst{23} \and Alexander Schlaefer\inst{12}}

\authorrunning{Nils Gessert et al.} 



\institute{$^1$Institute of Medical Technology, Hamburg University of Technology, Hamburg, Germany\\
$^2$DAISYlab, Forschungszentrum Medizintechnik Hamburg, Hamburg, Germany\\ 
$^3$Institute of Computational Neuroscience, University Medical Center Hamburg-Eppendorf, Hamburg, Germany \\
\email{nils.gessert@tuhh.de}}

%
%

\maketitle

\begin{abstract} 

In this paper, we describe our method for the ISIC 2019 Skin Lesion Classification Challenge. The challenge comes with two tasks. For task 1, skin lesions have to be classified based on dermoscopic images. For task 2, dermoscopic images and additional patient meta data have to be used. A diverse dataset of \num{25000} images was provided for training, containing images from eight classes. The final test set contains an additional, unknown class. We address this challenging problem with a simple, data driven approach by including external data with skin lesions types that are not present in the training set. Furthermore, multi-class skin lesion classification comes with the problem of severe class imbalance. We try to overcome this problem by using loss balancing. Also, the dataset contains images with very different resolutions. We take care of this property by considering different model input resolutions and different cropping strategies. To incorporate meta data such as age, anatomical site, and sex, we use an additional dense neural network and fuse its features with the CNN. We aggregate all our models with an ensembling strategy where we search for the optimal subset of models. Our best ensemble achieves a balanced accuracy of $\SI{74.2}{\percent}$ using five-fold cross-validation. On the official test set our method is ranked first for both tasks with a balanced accuracy of $\SI{63.6}{\percent}$ for task 1 and $\SI{63.4}{\percent}$ for task 2.

\end{abstract}

\section{Introduction}

Automated skin lesion classification is a challenging problem that is typically addressed using convolutional neural networks. Recently, the ISIC 2018 Skin Lesion Analysis Towards Melanoma Detection challenge resulted in numerous high-performing methods that performed similar to human experts for evaluation of dermoscopic images \cite{tschandl2019comparison}. To improve diagnostic performance further, the ISIC 2019 challenge comes with several old and new problems to consider. In particular, the test set of the ISIC 2019 challenge contains an unknown class that is not present in the dataset. Also, the severe class imbalance of real-world datasets is still a major point that needs to be addressed. Furthermore, the training dataset, previously HAM10000 \cite{tschandl2018ham10000}, was extended by additional data from the BCN\_20000 \cite{combalia2019bcn20000} and MSK dataset \cite{codella2018skin}. The images have different resolutions and were created using different preprocessing and preparation protocols that need to be taken into account. 

In this paper we describe our procedure for our participation in the two tasks of the ISIC 2019 Challenge. For task 1, skin lesions have to be classified based on dermoscopic images only. For task 2, dermoscopic images and additional patient meta data have to be used. We largely rely on established methods for skin lesion classifcation including loss balancing, heavy data augmentation, pretrained, state-of-the-art CNNs and extensive ensembling \cite{gessert2018skin,gessert2019skin}. We address data variability by applying a color constancy algorithm and a cropping algorithm to deal with raw, uncropped dermoscopy images. We deal with the unknown class in the test set with a data driven approach by using external data. For task 2, we incorporate additional meta information into the model using a dense neural network which is fused with the CNN's features.

\section{Materials and Methods}

\subsection{Datasets}

The main traning dataset contains \num{25331} dermoscopic images, acquired at multiple sites and with different preprocessing methods applied beforehand. It contains images of the classes melanoma (MEL), melanocytic nevus (NV), basal cell carcinoma (BCC), actinic keratosis (AK), benign keratosis (BKL), dermatofibroma (DF), vascular lesion (VASC) and squamous cell carcinoma (SCC). A part of the training dataset is the HAM10000 dataset which contains images of size $600\times 450$ that were centered and cropped around the lesion. The dataset curators applied histogram corrections to some images \cite{tschandl2018ham10000}. Another dataset, BCN\_20000, contains images of size $1024\times 1024$. This dataset is particularly challenging as many images are uncropped and lesions in difficult and uncommon locations are present \cite{combalia2019bcn20000}. Last, the MSK dataset contains images with various sizes.

The dataset also contains meta information about the patient's age group (in steps of five years), the anatomical site (eight possible sites) and the sex (male/female). The meta data is partially incomplete, i.e., there are missing values for some images.

In addition, we make use of external data. We use the \num{995} dermoscopic images from the 7-point dataset \cite{kawahara20187}. Moreover, we use an in-house dataset which consists of \num{1339} images. The in-house dataset also contains \num{353} images that we use for the unknown class (UNK). We include images of healthy skin, angiomas, warts, cysts and other benign alterations. The key idea is to build a broad class of skin variations which should encourage the model to assign any image that is not part of the eight main classes to the ninth broad pool of skin alterations. We also consider the three types of meta data for our external data, if it is available. 

For internal evaluation we split the main training dataset into five folds. The dataset contains multiple images of the same lesions. Thus, we ensure that all images of the same lesion are in the same fold. We add all our external data to each of the training sets. Note that we do not include any of our images from the unknown class in our evaluation as we do not know whether they accurately represent the actual unknown class. Thus, all our models are trained to predict nine classes but we only evaluate on the known, eight classes.

We use the mean sensitivity for our internal evaluation which is defined as

\begin{equation}
	S = \frac{1}{C}\sum_{i=1}^{C}\frac{\mathit{TP}_i}{\mathit{TP}_i+\mathit{FN}_i}
\end{equation}

where $\mathit{TP}$ are true positives, $\mathit{FN}$ are false negatives and $C$ is the number of classes. The metric is also used for the final challenge ranking.

\subsection{Preprocessing}

\begin{figure}[ht]
\begin{center}
   \includegraphics[width=0.32\linewidth]{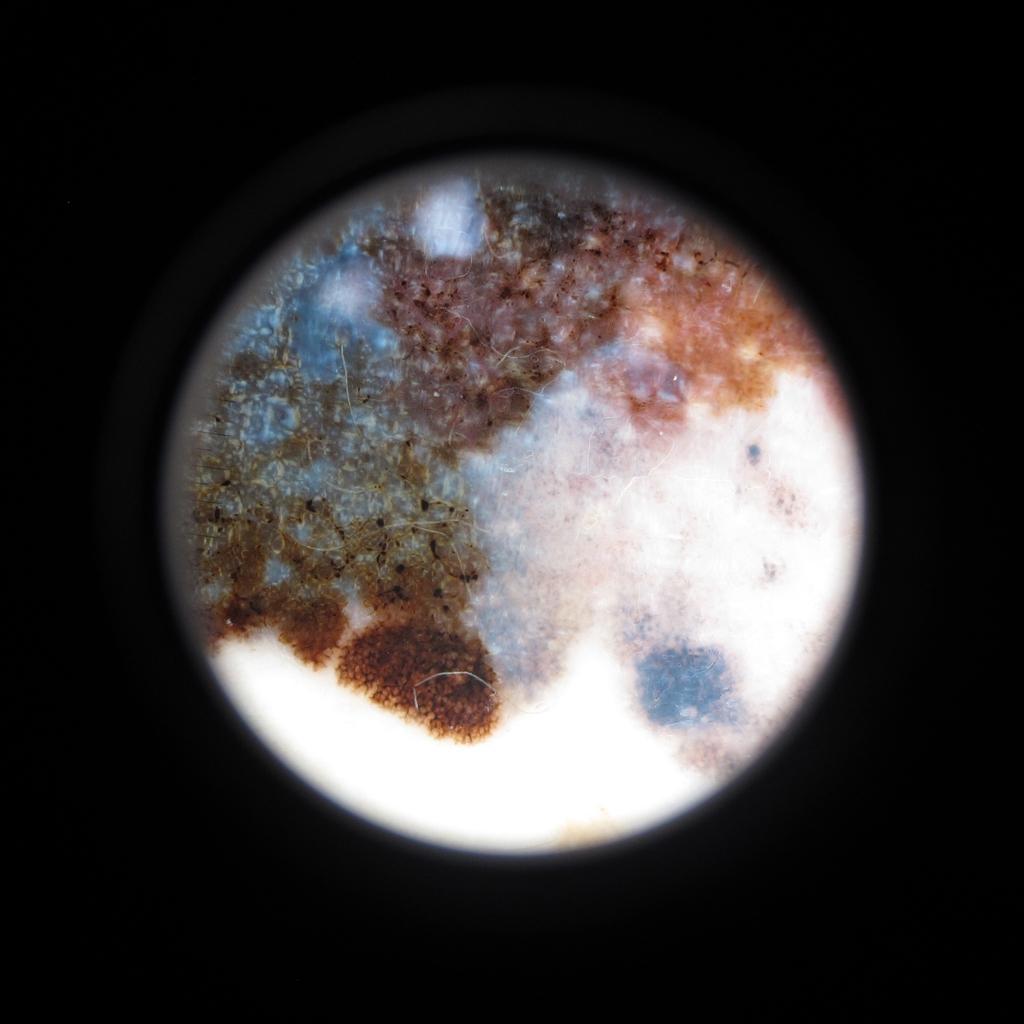}
   \includegraphics[width=0.32\linewidth]{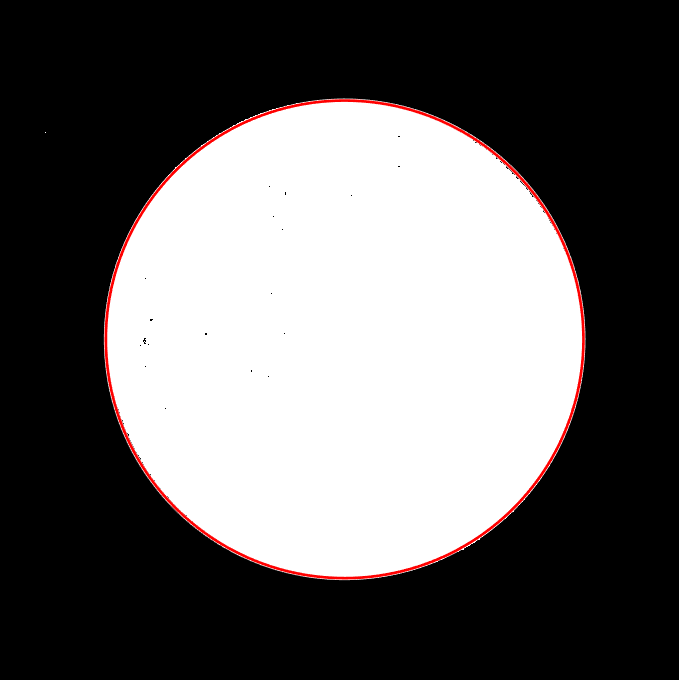}
   \includegraphics[width=0.32\linewidth]{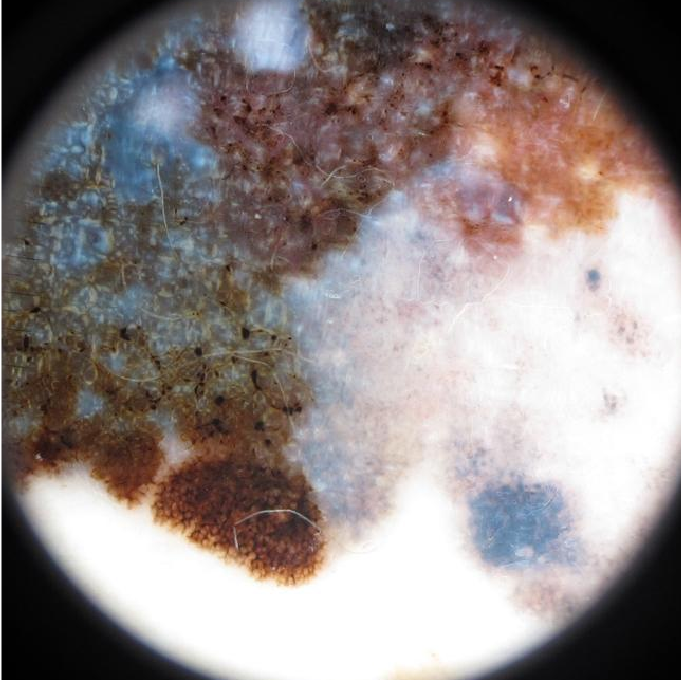}
\end{center}
   \caption{Our cropping strategy for uncropped dermoscopy images. Left, the original images is shown. In the center, the binarized image with the circular fit is shown. Right, the cropped image is shown.} 
\label{fig:cropping}
\end{figure}

\textbf{Image Preprocessing.} As a first step, we use a cropping strategy to deal with the uncropped images which often show large, black areas. We binarize the images with a very low threshold, such that the entire dermoscopy field of view is set to \num{1}. Then, we find the center of mass and the major and minor axis of an ellipse that has the same second central moments as the inner area. Based on these values we derive a rectangular bounding box for cropping that covers the relevant field of view. The processes is illustrated in Figure~\ref{fig:cropping}. We automatically determin the necessecity for cropping based on a heuristic that tests whether the mean intensity inside the bounding box is substantially different from the mean intensity outside of the bounding box. Manual inspection showed that the method was robust. In the training set, \num{6226} were automatically cropped. In the test set, \num{3864} images were automatically cropped. Next, we apply the Shades of Gray color constancy method with Minkowski norm $p=6$, following last year's winner \cite{codella2019skin}. This is particular important as the datasets used for training differ a lot. Furthermore, we resize the larger images in the datasets. We take the HAM10000 resolution as a reference and resize all images' longer side to \num{600} pixels while preserving the aspect ratio. 

\textbf{Meta Data Preprocessing.} For task 2, our approach is to use the meta data with a dense (fully-connected) neural network. Thus, we need to encode the data as a feature vector. For the anatomical site and sex, we chose a one-hot encoding. Thus, the anatomical site is represented by eight features where one of those features is one and the others are zero for each lesion. The same applies for sex. In case the value is missing, all features for that property are zero. For age, we use a normal, numerical encoding, i.e. age is represented by a single feature. This makes encoding missing values difficult, as the missingness should not have any meaning (we assume that all values are missing at random). We encode a missing value as $\num{-5}$ as $\num{0}$ is also part of the training set's value range. To overcome the issue of missing value encoding, we also considered a one-hot encoding for the age groups. However, initial validation experiments should slightly worse performance which is why we continued with the numerical encoding.

\subsection{Deep Learning Models} \label{sec:model}

\begin{figure}[ht]
\begin{center}
   \includegraphics[width=1.0\linewidth]{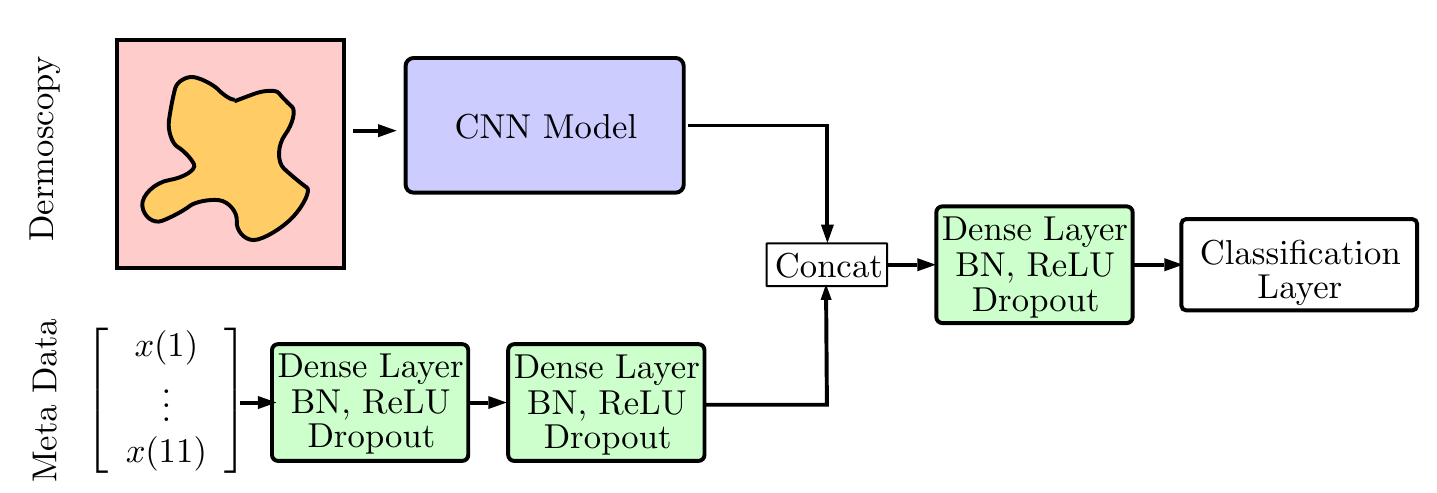}
\end{center}
   \caption{Approach for adding meta data to a CNN model. BN refers to batch normalization.} 
\label{fig:approach}
\end{figure}

\textbf{General Approach.} For task 1, we employ various CNNs for classifying dermoscopic images. For task 2, our deep learning models consist of two parts, a CNN for dermoscopy images and a dense neural network for meta data. The approach is illustrated in Figure~\ref{fig:approach}. As a first step, we train our CNNs on image data only (task 1). Then, we freeze the CNNs weights and attach the meta data neural network. In the second step, we only train the meta data network's weights and the classification layer. We describe CNN training first, followed by the meta data training.

\textbf{CNN Architectures.} We largely rely on EfficientNets (EN) \cite{tan2019efficientnet} that have been pretrained on the ImageNet dataset with the AutoAugment v0 policy \cite{cubuk2018autoaugment}. This model family contains eight different models that are structurally similar and follow certain scaling rules for adjustment to larger image sizes. The smallest version B0 uses the standard input size $224\times 224$. Larger versions, up to B7, use increased input size while also scaling up network width (number of feature maps per layer) and network depth (number of layers). We employ EN B0 up to B6. For more variability in our final ensemble, we also include a SENet154 \cite{hu2018squeeze} and two ResNext variants pretrained with weakly supervised learning (WSL) on \num{940} million images \cite{mahajan2018exploring}.

\textbf{CNN Data Augmentation.} Before feeding the images to the networks, we perform extensive data augmentation. We use random brightness and contrast changes, random flipping, random rotation, random scaling (with appropriate padding/cropping), and random shear. Furthermore, we use CutOut \cite{devries2017improved} with one hole and a hole size of \num{16}. We tried to apply the AutoAugment v0 policy, however, we did not observe better performance.

\textbf{CNN Input Strategy.} We follow different input strategies for training that transform the images from their original size after preprocessing to a suitable input size. First, we follow a same-sized cropping strategy which we employed in the last year's challenge \cite{gessert2018skin}. Here, we take a random crop from the preprocessed image. Second, we follow a random-resize strategy which is popular for ImageNet training \cite{Szegedy.2016b}. Here, the image is randomly resized and scaled when taking a crop from the preprocessed image.

\textbf{CNN Training.} We train all models for \num{100} epochs using Adam. We use a weighted cross-entropy loss function where underrepresented classes receive a higher weight based frequency in the training set. Each class is multiplied by a factor $n_i = \left(N/N_i\right)^k$ where $N$ is the total number of training images, $N_i$ is the number of images in class $i$ and $k$ controls the balancing severity. We found $k=1$ to work best. We also tried to use the focal loss \cite{lin2017focal} with the same balancing weights without performance improvements. Batch size and learning rate are adopted based on GPU memory requirements of each architecture. We halve the learning every \num{25} epochs. We evaluate every \num{10} epochs and save the model achieving the best mean sensitivity (best). Also, we save the last model after \num{100} epochs of training (last). Training is performed on NVIDIA GTX 1080TI (B0-B4) and Titan RTX (B5,B6) graphics cards.

\textbf{Meta Data Architecture.} For task 2, the meta data is fed into a two-layer neural network with $256$ neurons each. Each layer contains batch normalization, a ReLU activation and dropout with $p=0.4$. The network's output is concatenated with the CNN's feature vector after global average pooling. Then, we apply another layer with batch normalization, ReLU and dropout. As a baseline we use $1024$ neurons which is scaled up for larger models, using EfficientNet's scaling rules for network width. Then, the classification layer follows. 

\textbf{Meta Data Augmentation.} We use a simply data augmentation strategy to address the problem of missing values. During training, we randomly encode each property as missing with a probability of $p=0.1$. We found this to be necessary as our images for the unknown class do not have any meta data. Thus, we need to ensure that our models do not associate missingness with this class.

\textbf{Meta Data Training.} During meta data training, the CNN's weights remain fixed. We still employ our CNN data augmentation strategies described above, i.e., the CNN still performs forward passes during training and the CNN's features are not fixed for each image. The meta data layers, i.e., the two-layer network, the layer after concatenation and the classification layer are trained for $\num{50}$ epochs with a learning rate of $l_r = \num{e-5}$ and a batch size of $\num{20}$.

\textbf{Prediction.} After training, we create predictions, depending on the CNN's input strategy. For same-sized cropping we take \num{36} ordered crops from the preprocessed image and average the softmaxed predictions of all crops. For random-resize cropping, we perform \num{16} predictions for each image with four differently scaled center crops and flipped versions of the preprocessed images. For the meta data, we  pass the same data through the network for each crop. Again, the softmaxed predictions are averaged.

\textbf{Ensembling.} Finally, we create a large ensemble out of all our trained models. We use a strategy where we select the optimal subset of models based on cross-validation performance \cite{gessert2019left}. Consider $C=\{c_1, \dots, c_n\}$ configurations where each configuration uses different hyperparameters (e.g. same-sized cropping) and baseline architectures (e.g. EN B0). Each configuration $c_i$ contains $m = 5$ trained models (best), one for each cross-validation split $v_j$. We obtain predictions $\hat{y}_{j}^{i}$ for each $c_i$ and $v_j$. Then, we perform an exhaustive search to find $C* \subseteq C$ such that $\hat{y}* = \frac{1}{|C*|}\sum_{i \in C*}\frac{1}{m}\sum_{j=1}^{m} \hat{y}_{j}^{i}$ maximizes the mean sensitivity $S$. We consider our $\num{8}$ top performing configurations from the ISIC 2019 Challenge Task 1 in terms of CV performance in $C$. We perform the search using the best models found during training only but but we also include the last models in the final ensemble to have a larger variability. Finally, we obtain predictions for the final test set using all models of all $c_i \in C*$. 

\section{Results}

\begin{table}[t]
\setlength{\tabcolsep}{10.5pt}
\caption{All cross-validation results for different configurations. We consider same-sized cropping (SS) and random-resize cropping (RR) and different model input resolutions. Values are given in percent as mean and standard deviation over all five CV folds. Ensemble average refers to averaging over all predictions from all models. Ensemble optimal refers to our search strategy for the optimal subset of configurations. $C=8$ refers to training with eight classes without the unknown class. T1 refers to Task 1 without meta data and T2 refers to Task 2 with meta data. ResNext WSL 1 and 2 refer to ResNeXt-101 WSL 32x8d and 32x16d, respectively \cite{mahajan2018exploring}.}
\begin{center}
\begin{tabular}{l l l}
\hline
\textbf{Configuration} & \textbf{Sensitivity T1} & \textbf{Sensitivity T2}  \\
\hline
 SENet154 SS $224\times 224$ & $67.2\pm 0.8$ & $70.0 \pm 0.8$ \\
 ResNext WSL 1 SS $224\times 224$ & $65.9\pm 1.6$ & $68.1 \pm 1.3$ \\
 ResNext WSL 2 SS $224\times 224$ & $65.3\pm 0.8$ & $69.1 \pm 1.5$ \\
 EN B0 SS $224\times 224$ $C=8$ & $66.7\pm 1.8$ & $68.8 \pm 1.5$ \\
 EN B0 SS $224\times 224$ & $65.8 \pm 1.7$ & $67.4 \pm 1.6$ \\
 EN B0 RR $224\times 224$ & $67.0 \pm 1.6$ & $68.9 \pm 1.7$ \\
 EN B1 SS $240\times 240$ & $65.9 \pm 1.6$ & $68.2 \pm 1.8$ \\ 
 EN B1 RR $240\times 240$ & $66.8 \pm 1.5$ & $68.5 \pm 1.8$ \\
 EN B2 SS $260\times 260$ & $67.2 \pm 1.4$ & $69.0 \pm 2.5$ \\ 
 EN B2 RR $260\times 260$ & $67.6 \pm 2.0$ & $70.1 \pm 2.0$ \\ 
 EN B3 SS $300\times 300$ & $67.8 \pm 2.0$ & $68.5 \pm 1.7$ \\ 
 EN B3 RR $300\times 300$ & $67.0 \pm 1.5$ & $68.4 \pm 1.5$ \\  
 EN B4 SS $380\times 380$ & $67.8 \pm 1.1$ & $68.5 \pm 1.1$ \\ 
 EN B4 RR $380\times 380$ & $68.1 \pm 1.6$ & $69.4 \pm 2.2$ \\ 
 EN B5 SS $456\times 456$ & $68.2 \pm 0.9$ & $68.7 \pm 1.6$ \\ 
 EN B5 RR $456\times 456$ & $68.0 \pm 2.2$ & $69.0 \pm 1.6$ \\   
 EN B6 SS $528\times 528$ & $68.8 \pm 0.7$ & $69.0 \pm 1.4$ \\ 
 Ensemble Average & $71.7 \pm 1.7$ & $73.4 \pm 1.6$ \\ 
 Ensemble Optimal & $72.5 \pm 1.7$ & $\bm{74.2 \pm 1.1}$ \\ 
 Official Testset & $63.6$ & $63.4$ \\  
 \hline
\end{tabular}
\end{center}
\label{tab:res}
\end{table}

For evaluation we consider the mean sensitivity $S$ for training with images only and for training with additional meta data. The results for cross-validation with individual models and our ensemble are shown in Table~\ref{tab:res}. Overall, large ENs tend to perform better. Comparing our input strategies, both appear to perform similar for most cases. Including the ninth class with different skin alterations slightly reduces performance for the first eight classes. Ensembling leads to substantially improved performance. Our optimal ensembling strategy improves performance slightly. The optimal ensemble contains nine out of the sixteen configurations.

Regarding meta data, performance tends to improve by $1$ to $2$ percent points through the incorporation of meta data. This increase is mostly observed for smaller models as larger models' show only minor performance changes. The final ensemble shows improved performance.

For our final submission to the ISIC 2019 Challenge task 1 we created an ensemble with both the best and last model checkpoints. For task 2, we submitted an ensemble with the best model checkpoints only and an ensemble with both best and last model checkpoints. The submission with only the best model checkpoints performed better. Overall, the performance on the official test set is substantially lower than the cross-validation performance. The performance for task2 is lower than the performance for task 1.

\begin{table}[t]
\setlength{\tabcolsep}{6.5pt}
\caption{Results from the official test set of the ISIC 2019 Challenge for each class. We consider the AUC, the AUC for a sensitivity larger than $\SI{80}{\percent}$ (AUC-S), the sensitivity and specificity. Note that the sensitivity given here is differently calculated than $S$. Values are given in percent.}
\begin{center}
\begin{tabular}{l l l l l l l l l l}
\hline
\textbf{Class} & \multicolumn{4}{c}{\textbf{Task 1}} & & \multicolumn{4}{c}{\textbf{Task 2}}  \\
 & AUC & AUC-S & Sens. & Spec. & & AUC & AUC-S & Sens. & Spec. \\
\hline
MEL & $92.8$ & $84.9$ & $59.4$ & $96.2$ & & $93.1$ & $84.9$ & $54.5$ & $97.6$ \\
NV & $96.0$ & $93.0$ & $71.0$ & $97.5$ & & $96.0$ & $93.2$ & $63.7$ & $98.3$ \\
BCC & $94.9$ & $90.4$ & $72.1$ & $94.0$ & & $94.7$ & $90.1$ & $64.9$ & $95.8$ \\
AK & $91.4$ & $82.4$ & $48.4$ & $96.5$ & & $91.9$ & $84.1$ & $46.0$ & $96.6$ \\
BKL & $90.4$ & $80.5$ & $39.4$ & $98.5$ & & $90.8$ & $82.1$ & $32.4$ & $99.1$ \\
DF & $97.9$ & $96.3$ & $57.8$ & $99.2$ & & $98.0$ & $96.5$ & $55.6$ & $99.3$ \\
VASC & $95.6$ & $92.5$ & $64.4$ & $99.1$ & & $94.2$ & $91.2$ & $49.5$ & $99.5$ \\
SCC & $93.8$ & $87.6$ & $43.9$ & $98.6$ & & $93.0$ & $87.8$ & $40.8$ & $98.7$ \\
UNK & $77.5$ & $58.1$ & $0.3$ & $99.9$ & & $61.2$ & $25.3$ & $0.0$ & $99.9$ \\
 \hline
\end{tabular}
\end{center}
\label{tab:res_official}
\end{table}

Table~\ref{tab:res_official} shows several metrics for the performance on the official test set. For task 1, the performance for the unknown class is substantially lower than for all other classes across several metrics. For task 2, the performance for the unknown class is also substantially reduced, compared to task 1.

\section{Discussion}

We explore multi-resolution EfficientNets for skin lesion classification, combined with extensive data augmentation, loss balancing and ensembling for our participation in the ISIC 2019 Challenge. In previous challenges, data augmentation and ensembling were key factors for high-performing methods \cite{codella2019skin}. Also, class balancing has been studied \cite{gessert2019skin} where loss weighting with a cross-entropy loss functions performed very well. We incorporate this prior knowledge in our approach and also consider the input resolutions as an important parameter. Our results indicate that models with a large input size perform better, see Table~\ref{tab:res}. For a long time, small input sizes have been popular and the effectiveness of an increased input size is likely tied to EfficientNet's new scaling rules \cite{tan2019efficientnet}. EfficientNet scales the models' width and depth according to the associated input size which lead to high-performing models with substantially lower computational effort and fewer parameters compared to other methods. We find that these concepts appear to transfer well to the problem of skin lesion classification. 

When adding meta data to the model, performance tends to improve slightly for our cross-validation experiments. The improvement is particularly large for smaller, lower-performing models. This might indicate that meta data helps models that do not leverage the full information that is available in the images alone.

The ISIC 2019 Challenge also includes the problem to predict an additional, unknown class. At the point of submission, there was no labeled data available for the class, thus, cross-validation results do not reflect our model's performance with respect to this class. The performance on the official test provides some insights into the unknown class, see Table~\ref{tab:res_official}. First, it is clear that the performance on the unknown class is substantially lower than the performance on the other classes. This could explain why there is a substantial difference between our cross-validation results and the results on the official test set. Second, we can observe a substantial performance reduction for the unknown class between task 1 and task 2. This might explain the lack of improvement for task 2, although our cross-validation performance improved with additional meta data. This is likely linked to the fact that we do not have meta data for our unknown class training images. Although we tried to overcome the problem with meta data dropout, our models appear to overfit to the characteristic of missing data for the unknown class.

Overall, we find that EfficientNets perform well for skin lesion classification. In our final ensembling strategy, various EfficientNets were present, although the largest ones performed best. This indicates that a mixture of input resolutions is a good choice to cover multi-scale context for skin lesion classification. Also, SENet154 and the ResNext models were automatically selected for the final ensemble which indicates that some variability in terms of architectures is helpful. 

\section{Conclusion}

In this paper we describe our method that ranked first in both tasks of the ISIC 2019 Skin Lesion Classification Challenge. We overcome the typical problem of severe class imbalance for skin lesion classification with a loss balancing approach. To deal with multiple image resolutions, we employ various EfficientNets  with different input cropping strategies and input resolutions. For the unknown class in the challenge's test set, we take a data driven approach with images of healthy skin. We incorporate meta data into our model with a two-path architecture that fuses both dermoscopic and meta data information. While setting a new state-of-the-art for skin lesion classification, we find that predicting an unknown class and the optimal use of meta data are still challenging problems. 

\bibliographystyle{spmpsci}      
\bibliography{egbib} 

\end{document}